\documentclass{bmvc2k}

\title{Temporal Lidar Depth Completion}

\addauthor{Pietari Kaskela}{pkaskela@nvidia.com}{1}
\addauthor{Philipp Fischer}{pfischer@nvidia.com}{1}
\addauthor{Timo Roman}{troman@nvidia.com}{1}

\addinstitution{
 Applied Deep Learning Research\\
 NVIDIA
}

\runninghead{Kaskela, Fischer, Roman}{Temporal Lidar Depth Completion}

\usepackage{enumitem}  
\usepackage{tikz}
\newcommand{\cspngrid}{\tikz{\draw[step=0.2,black,thin] (0,0) grid (0.6,0.6);}}
\usetikzlibrary{positioning, arrows.meta, shapes, shadows, matrix}
\usepackage{graphicx}
\usepackage{amsfonts,amssymb,amsbsy,amsmath}
\usepackage{gensymb}
\usepackage{float}
\usepackage{booktabs}
\usepackage{algorithm}
\usepackage{algpseudocode}
\usepackage{caption}
\usepackage{adjustbox}
\usepackage{wrapfig}

\newcommand*\annotatedFigureBoxCustom[8]{\draw[#5,thick,rounded corners] (#1) rectangle (#2);\node at (#4) [fill=#6,thick,shape=circle,draw=#7,inner sep=1pt,font=\sffamily, font=\small,text=#8] {\textbf{#3}};}
\newcommand*\annotatedLabel[2]{\node at (#1) [fill=white,thick,shape=rectangle,draw=black,inner sep=2pt,font=\sffamily,font=\small, text=black, anchor=south west] {\textbf{#2}};}
\newcommand*\annotatedFigureBox[4]{\annotatedFigureBoxCustom{#1}{#2}{#3}{#4}{white}{white}{black}{black}}

\newenvironment {annotatedFigure}[1]{\centering\begin{tikzpicture}
    \node[anchor=south west,inner sep=0] (image) at (0,0) { #1};\begin{scope}[x={(image.south east)},y={(image.north west)}]}{\end{scope}\end{tikzpicture}}
\begin{document}

\maketitle
\begin{abstract}
Given the lidar measurements from an autonomous vehicle, we can project the points and generate a sparse depth image. Depth completion aims at increasing the resolution of such a depth image by infilling and interpolating the sparse depth values.

Like most existing approaches, we make use of camera images as guidance in very sparse or occluded regions. In addition, we propose a temporal algorithm that utilizes information from previous timesteps using recurrence. In this work, we show how a state-of-the-art method PENet can be modified to benefit from recurrency. Our algorithm achieves state-of-the-art results on the KITTI depth completion dataset while adding only less than one percent of additional overhead in terms of both neural network parameters and floating point operations. The accuracy is especially improved for faraway objects and regions containing a low amount of lidar depth samples. Even in regions without any ground truth (like sky and rooftops) we observe large improvements which are not captured by the existing evaluation metrics.
\end{abstract}

\section{Introduction}

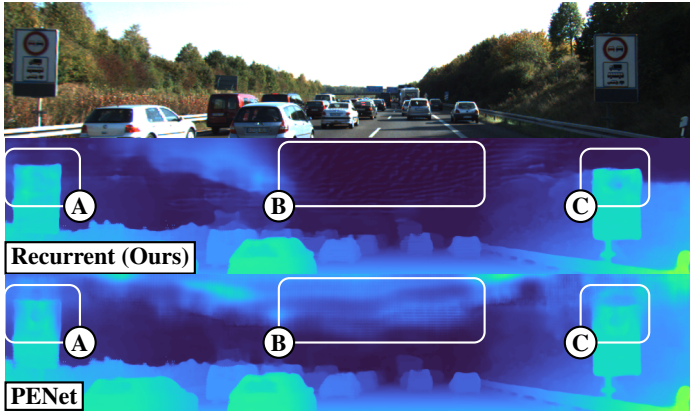
\begin{figure}[t]
\setlength{\lineskip}{0pt}
\centering
    \includegraphics[trim={0cm 5cm 8cm 0.5cm},clip,width=0.7\textwidth]{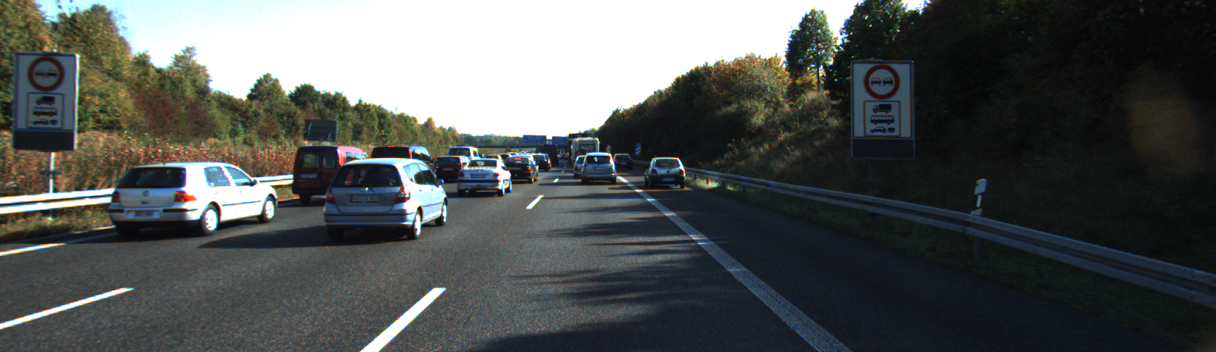}\\
     \begin{annotatedFigure}%
     {\includegraphics[trim={0cm 5cm 8cm 0.5cm},clip,width=0.7\textwidth]{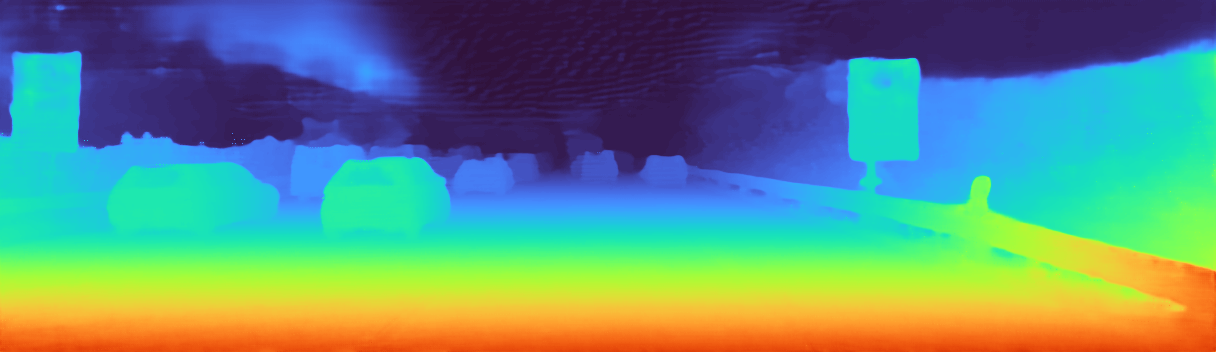}}
     \annotatedFigureBox{0.84,0.5}{0.94,0.92}{C}{0.84,0.5}%bl
     \annotatedFigureBox{0.4,0.5}{0.7,0.97}{B}{0.4,0.5}%bl
     \annotatedFigureBox{0.0,0.5}{0.11,0.92}{A}{0.11,0.5}%bl
     \annotatedLabel{0, 0}{Recurrent (Ours)}
     \end{annotatedFigure}\\
     \begin{annotatedFigure}
     {\includegraphics[trim={0cm 5cm 8cm 0.5cm},clip,width=0.7\textwidth]{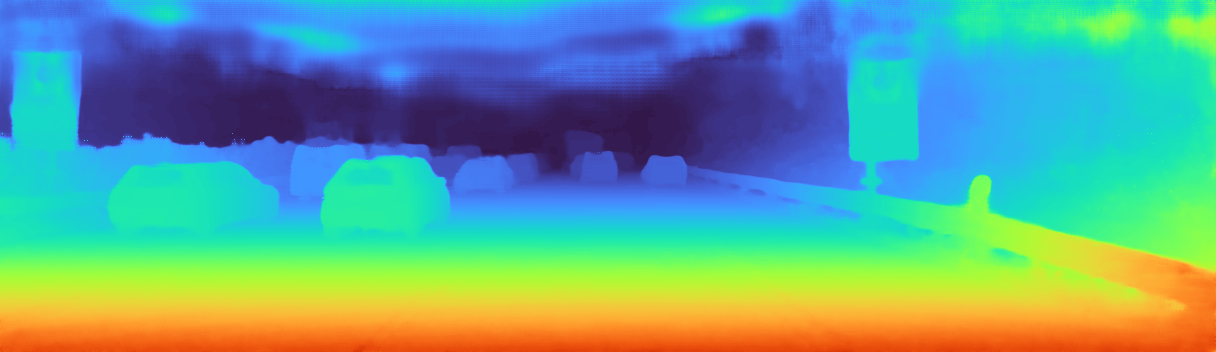}}
     \annotatedFigureBox{0.84,0.5}{0.94,0.92}{C}{0.84,0.5}%bl
     \annotatedFigureBox{0.4,0.5}{0.7,0.97}{B}{0.4,0.5}%bl
     \annotatedFigureBox{0.0,0.5}{0.11,0.92}{A}{0.11,0.5}%bl
     \annotatedLabel{0, 0}{PENet}
     \end{annotatedFigure}\\
     \vspace*{0.3cm}
     \caption{Our recurrent model achieves a new state-of-the-art result on the KITTI depth completion validation set. In addition, our model significantly improves upon regions which are not captured by the sparse ground truth or lidar input, as illustrated in regions (A), (B) and (C).}
     \label{fig:sota_vs_penet2}
\end{figure}
\begin{figure}[tb]
\centering
     \resizebox{\textwidth}{!}{\input{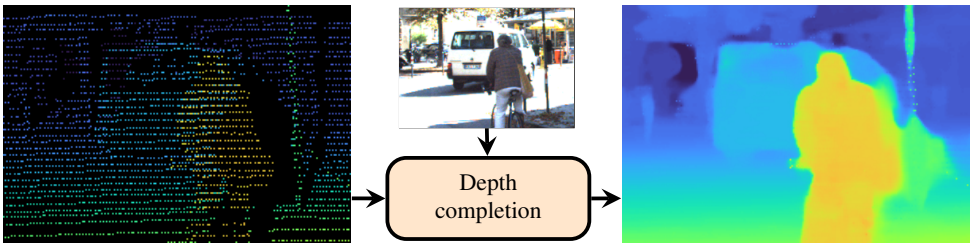}}\\
     \vspace*{0.3cm}
\caption{A cropped color image (middle), with the corresponding projected sparse lidar sample (left) and a depth predicted by a depth completion model (right). Notice how the projected sparse lidar samples for the cyclist and the van overlap because of the different viewpoints of the sensors. The camera is mounted to the right of the lidar sensor.}
\label{fig:in_out_crop}
\end{figure}

The task of depth completion aims at recovering a dense depth map from a sparse depth map using additional inputs such as camera images as guidance (cf. \autoref{fig:in_out_crop}). The task is especially important in the context of autonomous vehicles (AVs), where sparse depth maps are produced by lidar sensors but dense depth maps are required by some employed perception algorithms. For example, the Velodyne HDL-64E lidar sensor used by the popular KITTI \cite{Geiger2013IJRR} dataset fills up only 6\% of the depth values of a corresponding color image, when projected onto the image.

In addition to infilling and interpolating the depth values of the remaining 94\% pixels, a proper depth completion solution needs to be able to deal with errors caused by the different mounting positions of the camera and lidar sensor, moving objects and the spinning movement of the lidar sensor itself. \autoref{fig:in_out_crop} illustrates the inputs (color image, sparse depth) and the output (dense depth) of the depth completion task. Notice how there are occlusions (image regions with missing points) and overlaps (image regions with points from different depths) in the image, since the lidar and the camera have slightly different viewpoints.

Most state-of-the-art depth completion approaches rely on a U-Net \cite{ronneberger2015unet} style backbone followed by a CSPN-based \cite{cheng2018depth} refinement network \cite{hu2020PENet, 9918022, lin2022dynamic}. For the closely related \textit{depth estimation} task, in which only camera images are available, temporal techniques have been extensively studied \cite{wang2019recurrent,8575528,rec_dc3,zhang2019exploiting}. Prior work on temporal techniques in the depth completion setting can be found in \cite{rec_dc3,rec_dc1,rec_dc2}, but these methods have not shown competitive accuracy with state-of-the-art non-temporal approaches.

In this paper, we propose a depth completion architecture to utilize temporal information for more accurate depth completion. Given any existing U-Net + refinement type approach, such as PENet \cite{hu2020PENet}, only minor modifications to the the number of input and output channels are required to implement our proposed temporal processing. We can further significantly improve the qualitative and quantitative results by utilizing pose information between timesteps to align the previous depth prediction with the next frame. 

Applying these improvements to the open-source PENet architecture allows us to achieve state-of-the-art results on the KITTI depth completion validation set, with a neglible increase in both model parameters and required floating point operations. In addition, our model significantly improves the depth completion accuracy in regions not captured by the sparse depth ground truth, as shown in \autoref{fig:sota_vs_penet2}. In summary, our contributions are as follows:
\begin{enumerate}[label=(\roman*)]
\item We present general modifications to PENet \cite{hu2020PENet} - a popular depth completion model - that allow the network to utilize temporal information. These modifications can be applied to other similar architectures.
\item We present an effective way of using pose information to warp the previous depth prediction to align with the current timestep and thereby significantly improve the accuracy.
\item Our depth completion solution achieves state-of-the-art accuracy on the KITTI depth completion dataset with a negligible increase in both neural network parameters and floating point operations required.
\end{enumerate}

\section{Related Work}
We rely on the KITTI depth completion dataset by \citet{uhrig2017sparsity}, a popular dataset for depth completion in the AV setting, which is an extension to the KITTI raw dataset \cite{Geiger2013IJRR}. The raw KITTI dataset contains driving sessions recorded with an extensive sensor suite, of which relevant to this work are the RGB cameras, the lidar sensor, the global positioning system (GPS) sensor and the inertial measurement unit (IMU) sensor. In addition, the KITTI depth completion dataset contains 93k sparse depth maps and the corresponding generated semi-dense ground truth depth maps, raw point clouds and RGB images. \citet{uhrig2017sparsity} also present a sparse convolutional neural network architecture for depth completion based on the sparse lidar depth input only.

Modern approaches focus on image-guided depth completion, in which the sparse depth is projected to the viewpoint of a corresponding color image. Most state-of-the-art solutions \cite{9918022,hu2020PENet,lin2022dynamic} rely on a U-Net-like backbone network to extract a coarse depth estimate and fused color-depth features, and then feed both to a CSPN-based \cite{cheng2018depth} refinement network. Recently, transformer-based solutions \cite{zhangcompletionformer} have also reached competitive accuracies in the KITTI depth completion challenge.

Previous work \cite{rec_dc3,rec_dc1,rec_dc2} has shown that temporal information can be used to improve depth completion solutions, but those have not achieved better accuracy compared to state-of-the-art non-temporal models. \citet{rec_dc3} built a joint framework for depth estimation and depth completion based on the ConvLSTM architecture \cite{NIPS2015_07563a3f}. \citet{rec_dc1} explored integrating ConvGRU \cite{ballas2016delving} and three-dimensional convolutional layers into existing depth completion architectures. \citet{rec_dc2} aim to not only complete sparse depth inputs using variational recurrent neural networks, but to also predict future dense depth maps.

Temporal approaches are much more popular in the image-only depth estimation setting, where the ill-posed nature of estimating 3-D depth from 2-D images alone is alleviated by exploiting the temporal correlations. Structure from Motion (SfM) \cite{sfm} techniques have been studied for decades. Effective ways of utilizing the temporal information in modern approaches include implementing a recurrent architecture such as in \cite{wang2019recurrent,8575528,rec_dc3,zhang2019exploiting} or utilizing the sequences to calculate consistency losses between frames \cite{zhou2017unsupervised,wang2019recurrent,godard2019digging}.

We chose to build upon the PENet \cite{hu2020PENet} model, as it is open-source and shares a similar structure with many other state-of-the-art depth completion solutions. The architecture consists of a backbone with two U-Nets and a dilated and accelerated convolutional spatial propagation network (DA-CSPN++) refinement network. The backbone takes as input the sparse lidar depth and the RGB image. The outputs of the backbone are a coarse depth estimate and internal features from the U-Nets. The coarse depth is then refined by the refinement module using an iterative procedure that propagates the nearby depth values based on the internal U-Net features and the sparse lidar depth.

\section{Method} \label{sec:temporal}

\begin{figure}[tb]
\centering
     \resizebox{\textwidth}{!}{\input{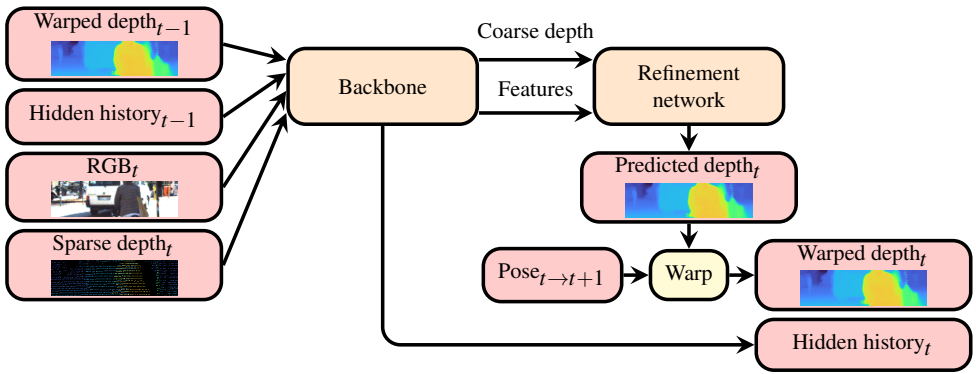}}\\
     \vspace*{0.3cm}
\caption{A general diagram of our recurrent depth completion model. The warped depth and hidden history from the previous timestep $t-1$ are fed as input to the backbone in the current timestep $t$.}
\label{fig:algo}
\end{figure}

To add recurrence to the PENet \cite{hu2020PENet} architecture, we chose to introduce additional input channels for the previous depth and hidden history as well as the corresponding output channel for the hidden history. If pose information between timesteps is available, we propose utilizing it to project the previous depth prediction to match with the current timestep. A general diagram of the whole model is illustrated in \autoref{fig:algo}.

\subsection{Architecture}

Since consecutive depth maps have a large correlation, we chose to base the recurrence of the model on the dense depth prediction of the previous timestep. We modify the PENet architecture consisting of a double U-Net backbone and a DA-CSPN++ refinement network such that the output depth of the DA-CSPN++ module is concatenated as an input channel to both U-Nets in the next timestep. This will allow the network to accumulate depth samples over time and to estimate depth in regions which contain a low amount of depth samples much better. Of course, the previous predicted dense depth is rather inaccurate in many scenarios, such as in settings with lots of moving objects or when the vehicle itself is moving fast. Solutions for alleviating the problems related to the movement of the vehicle itself are discussed in \autoref{sec:warping}.

To allow the network to more freely learn to pass information between timesteps, a single-channel hidden history is also added to the architecture. This is implemented as an additional output channel from the last layer of the second U-Net and concatenated to the input of both of the U-Nets in the next timestep. This way, the hidden history contains latent information from the deeper and spatially smaller layers, but can also pass high-resolution information from earlier layers. The values of the hidden history are clamped between -1 and 1 to stabilize training.

\subsection{Warping} \label{sec:warping}
Naively feeding the previous dense depth prediction as input to the next timestep is not optimal, due to the misalignment caused by the movement of the vehicle itself and moving objects in the environment. It is possible for the network to learn to correct this misalignment, but we found that effective utilization of previous depth values requires correcting as much of the misalignment as possible before the neural network. The problem of predicting the three-dimensional movement of dynamic objects in the environment is called scene flow estimation, but integrating such capability is out of the scope of this paper. However, the misalignment between timesteps caused by the movement of the vehicle itself can be corrected when the egomotion of the vehicle is available (i.e. its relative translation and rotation between the timesteps).

We propose applying classic reprojection to the predicted depth to transform the depth to the viewpoint of the vehicle in the next timestep. The warping requires the relative pose matrix $P_{(t-1) \rightarrow t}$ describing the translation and rotation between the timesteps and the camera intrinsics matrix $K$ used for projecting the point-cloud to the image-plane and back. Possible overlaps caused by the reprojection are solved by choosing the minimum depth value at each pixel. Warping can be implemented in an efficient and differentiable way by utilizing the scatter-function available in modern deep learning frameworks. This work considered warping the previous depth only. One possible extension to this would be to study the impact of warping the hidden history in addition. This is left for further study.

For the KITTI depth completion dataset, we calculate the pose matrices from the GPS and IMU sensor data using the approach detailed in the original publication \cite{Geiger2013IJRR}. The pose matrices can also be estimated using visual odometry \cite{Geiger2012CVPR} solutions such as COLMAP \cite{schoenberger2016sfm}.

\subsection{Implementation details} 
The models were trained on a machine with either 8 NVIDIA
A100 GPUs, each with 40GB memory or 8 NVIDIA V100 GPUs, each with 32GB memory. The models were trained for up to 60 epochs using a cosine decay learning rate scheduler \cite{loshchilov2017sgdr} with a linear warmup of 2 epochs and a batch size of 4. The base learning rate was $10^{-3}$. All of the training runs used the AdamW optimizer \cite{loshchilov2019decoupled} with a weight decay of $10^{-6}$ and parameters $\beta_1 = 0.9$ and $\beta_2 = 0.99$. 

We train on the KITTI depth completion training set, which contains 86k images and provide results on the KITTI depth completion validation set which contains 4k images. The temporal models are trained on uniformly sampled sequences of 32 contiguous frames, while the validation uses sequences of length 128. For training, sequences with less than 32 frames are dropped, but validation is done on all frames of the validation set.

The training images are first bottom-center cropped to 352 by 1216 pixels to remove most of the sky, which does not contain any lidar samples, and then randomly cropped to the final training size of 192 by 608 pixels. The augmentations used during training include flipping the image and lidar inputs horizontally for 40\% of the sequences, color-jitter augmentation and dropping uniformly random-sized rectangles from the sparse lidar depth for 15\% of the frames. The image input is scaled to the interval [0, 1] and the depth input and ground truth values are divided by 100. The loss is the mean square error calculated against the valid ground truth pixels.

The networks are trained using truncated backpropagation through time (TBPTT) \cite{6797135}, which is parametrized by the weight update interval $k_1$ and the backpropagation length $k_2$. In \autoref{sec:exp} we provide an ablation on these parameters and show that temporally-aware training is required for optimal accuracy.

Our baseline non-temporal model and training procedure is derived from PENet \cite{hu2020PENet}. In addition to the previously mentioned changes related to training, we also removed the Batch Normalization \cite{ioffe2015batch} operations that were applied after the mask and kernel confidence generating convolutions in the DA-CSPN++ refinement network.

\section{Experiments} \label{sec:exp}
In this section we study the effects of the proposed architectural changes detailed in \autoref{sec:temporal} and compare our results to current state-of-the-art depth completion solutions. Following the KITTI depth completion benchmark, we report the following four metrics that compare the predicted depth $\hat{D}$ and the ground truth depth $D$: root mean square error (RMSE), mean absolute error (MAE), root mean square error of the inverse depth (iRMSE) and mean absolute error of the inverse depth (iMAE).

\subsection{Ablation study}
\autoref{tab:recurrent_results} compares our proposed architectural changes to a non-temporal baseline derived from PENet. We trained the baseline four times to gauge the variance between training runs. In the \textit{Previous Depth} -section of the table we introduce recurrence based on the unmodified depth output from the previous timestep and observe an absolute improvement of about 10 in the RMSE metric. The rather little gain is expected as the unwarped depth provides a useful guide to the network only in very limited static scenarios. 

\begin{table}[!htb]
\makebox[\textwidth][c]{
    \begin{tabular}{l | l | l | l | l}
    \midrule[1pt]
    \multicolumn{1}{p{3cm}|}{Configuration}&\multicolumn{1}{|p{1.3cm}|}{$\text{RMSE} \downarrow \newline \text{(mm)}$}&\multicolumn{1}{|p{0.9cm}|}{$\text{MAE} \downarrow$ \newline\text{(mm)}}&\multicolumn{1}{|p{0.9cm}|}{$\text{iRMSE} \downarrow$ \newline \text{(1/km)}}&\multicolumn{1}{|p{0.9cm}}{$\text{iMAE} \downarrow$ \newline\text{(1/km)}}\\ 
    \midrule[1pt]
    \textbf{Baseline} & 773.9$\pm$3.2 & 218.0$\pm$0.8 & 2.34$\pm$0.04 & 0.97$\pm$0.01\\
    \hline
    \textbf{Previous Depth} & & & &  \\
    \hline
    \hspace{0.3cm} + TBPTT(1, 1) & 762.4 (-11.5) & 215.1 & 2.24 & 0.94 \\
    \hline
    \hspace{0.3cm} + TBPTT(2, 2) & 772.0 (-1.9) & 216.9 & 2.29 & 0.96 \\
    \hline
    \hspace{0.3cm} + TBPTT(1, 2) + Hidden Hist. & 758.5 (-15.4) & 214.2 & 2.23 & 0.95\\
    \hline

    \textbf{Warped Previous Depth} & & & &  \\
    \hline
    \hspace{0.3cm} + TBPTT(1, 1) & 728.7 (-45.2) & 204.9 & 2.20 & 0.94 \\
    \hline
\hspace{0.3cm} + TBPTT(2, 2) & 737.4 (-36.5) & 209.3 & 2.31 & 0.97 \\
\hline
\hspace{0.3cm} + TBPTT(1, 2) + Hidden Hist. & \textbf{720.8 (-53.1)} & 203.5 & 2.25 & 0.94 \\
    \end{tabular}
}
\vspace{0.25cm}
\caption{Metrics for the full KITTI depth completion validation set for the recurrent experiments. The metrics are reported from the checkpoint of the network with the lowest RMSE metric. For the baseline experiment, mean and variance from four training runs are reported.} \label{tab:recurrent_results}
\end{table}

In the \textit{Warped Previous Depth} -section of \autoref{tab:recurrent_results} we introduce warping of the output depth between the timesteps, which results in significant improvement in all metrics. Combining the warped depth experiment with a single-channel hidden history and a temporally-aware training algorithm allows us to reach a RMSE of 720.8 on the full KITTI depth completion validation set.

As previously mentioned, we use truncated backpropagation through time as a training algorithm to explore the tradeoffs between the temporal capabilities of the model, the final accuracy and training time. The training algorithm TBPTT$(k_1,k_2)$ is parametrized by the weight update interval $k_1$ and the backpropagation length $k_2$. For faster training, we experimented on increasing the weight update interval parameter $k_1$ of TBPTT from one to two, but the value of one resulted in higher accuracy consistently in all of our experiments. To induce more temporal utilization at the cost of training time, we tried different values of the backpropagation length $k_2$. Backpropagation length of two is necessary for the network to be able to utilize the hidden history and based on our experiments also sufficient for realizing most of the temporal benefits.

\subsection{Comparison to state-of-the-art}
In \autoref{tab:sota_results} we compare our results to other state-of-the-art solutions on the KITTI depth completion validation set. On the KITTI depth completion validation set our results significantly improve upon the previous published state-of-the-art solution SemAttNet \cite{9918022}. Note that we are unable to provide results on the test set for our model as the test set is not published in a temporal format which is needed by our method. Unfortunately, the test set does not consist of consecutive frames.

The baseline PENet (containing a double U-Net backbone and the DA-CSPN++ refinement network) has about 130 million parameters and 404 GFLOPs. While significantly improving the results, our best recurrent model increases the parameter count by less than a hundredth of a percent and the FLOPs by 0.35\%. The overhead is small, since we only change the outermost layers of the neural networks. The additional depth warping overhead is negligible as well.

\begin{table}[!htb]
	\centering
    \begin{tabular}{l | l | l | l | l}
    \midrule[1pt]
    \multicolumn{1}{p{2cm}} {Model} & \multicolumn{1}{|p{1.5cm}|}{RMSE $\downarrow$ \newline \text{(mm)}} & \multicolumn{1}{p{1.5cm}}{MAE $\downarrow$ \newline \text{(mm)}} & \multicolumn{1}{|p{1.5cm}|}{$\text{iRMSE} \downarrow$ \newline \text{(1/km)}}&\multicolumn{1}{|p{1.5cm}}{$\text{iMAE} \downarrow$ \newline\text{(1/km)}} \\ 
    \midrule[1pt]
    \textbf{PENet \cite{hu2020PENet}} & 757.2  & 209.0 & 2.22 & 0.92\\
    \hline
    \textbf{RigNet \cite{DBLP:journals/corr/abs-2107-13802}} & 752.1 & 205.2 & 3.22 & 0.93\\
    \hline
    \textbf{DySPN \cite{lin2022dynamic}} & 739.4 & \textbf{191.4} & - & - \\
    \hline
    \textbf{SemAttNet \cite{9918022}} &  738.1 & 204.5 & 2.01 & 0.89 \\
    \hline
    \textbf{Recurrent (Ours)} & \textbf{722.2} & 204.0 & 2.30 & 0.96\\
    \end{tabular}
    \vspace{0.25cm}
    \caption{Metrics for the KITTI depth completion selected validation set with 1000 frames for several state-of-the-art methods. Our recurrent model outperforms all of the published models on the KITTI depth completion validation set. DySPN is the only method that trains on the L1-loss and thus has a significantly lower MAE.} \label{tab:sota_results}
\end{table} 

No additional training data was used by us except for the poses provided by the original KITTI dataset, compared to the other methods in \autoref{tab:sota_results}. Since the existing training set contains sequences, our recurrent method can use their order. This information (poses and frame sequences) is also available in a real-world
automotive setting, which we are targeting.

\subsection{Analysis}
As the semi-dense ground truth depth of the KITTI depth completion dataset still fills only 16\% of the pixels of the corresponding color image, standard metrics such as RMSE and MAE fail to capture differences between models in regions that contain a low amount of ground truth samples. 

\begin{figure}[!htb]
\setlength{\lineskip}{0pt}
\centering
    \includegraphics[trim={8cm 3cm 8cm 0},clip,width=0.4\textwidth]{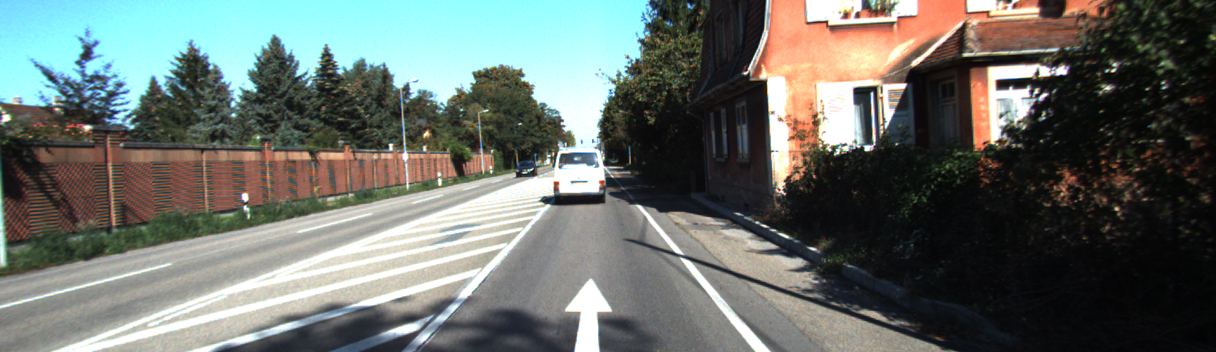}%
     \includegraphics[trim={8cm 3cm 8cm 0},clip,width=0.4\textwidth]{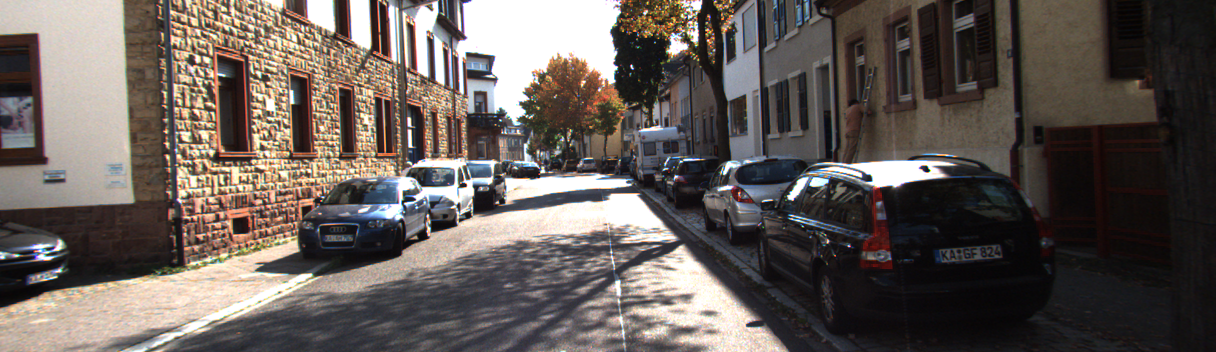}\\
     \begin{annotatedFigure}
    {\includegraphics[trim={8cm 3cm 8cm 0},clip,width=0.4\textwidth]{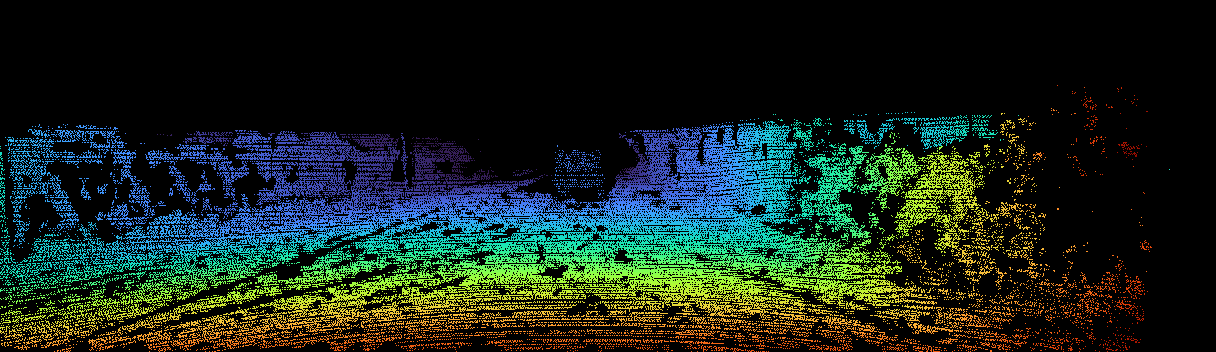}}
    \annotatedFigureBox{0.55,0.5}{0.8,0.98}{A}{0.8,0.5}%bl
     \annotatedLabel{0, 0}{Ground truth}
     \end{annotatedFigure}%
    \begin{annotatedFigure}
     {\includegraphics[trim={8cm 3cm 8cm 0},clip,width=0.4\textwidth]{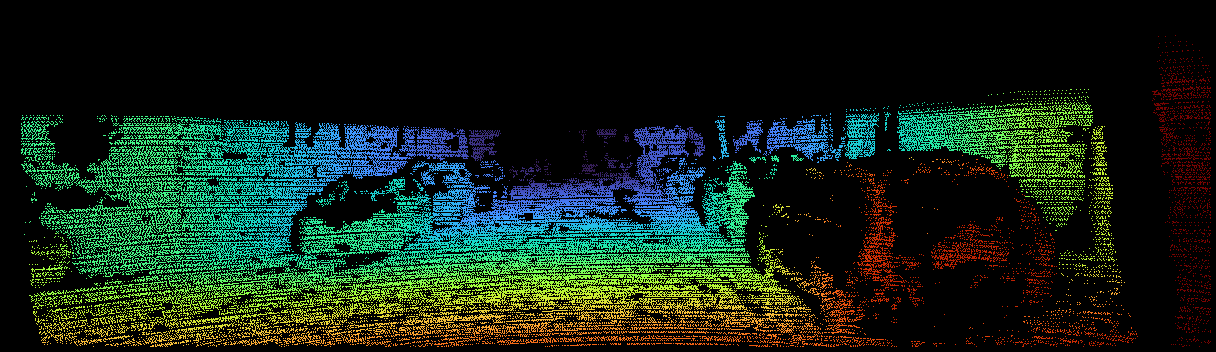}}
     \annotatedFigureBox{0.1,0.454}{0.35,0.974}{B}{0.1,0.454}%bl
     \annotatedFigureBox{0.56,0.67}{0.67,0.98}{C}{0.56,0.7}%bl
     \annotatedLabel{0, 0}{Ground truth}
     \end{annotatedFigure}\\
     \begin{annotatedFigure}%
     {\includegraphics[trim={8cm 3cm 8cm 0},clip,width=0.4\textwidth]{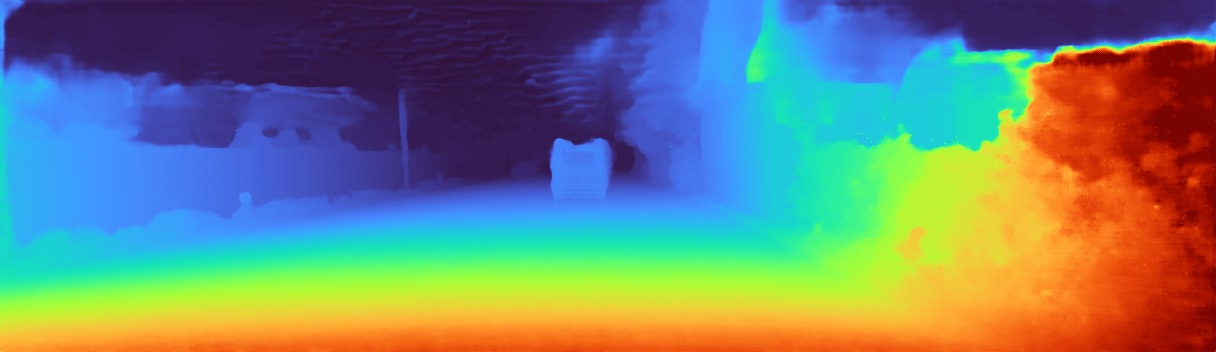}}
    \annotatedFigureBox{0.55,0.5}{0.8,0.98}{A}{0.8,0.5}%bl
     \annotatedLabel{0, 0}{Recurrent (Ours)}
     \end{annotatedFigure}%
     \begin{annotatedFigure}
     {\includegraphics[trim={8cm 3cm 8cm 0},clip,width=0.4\textwidth]{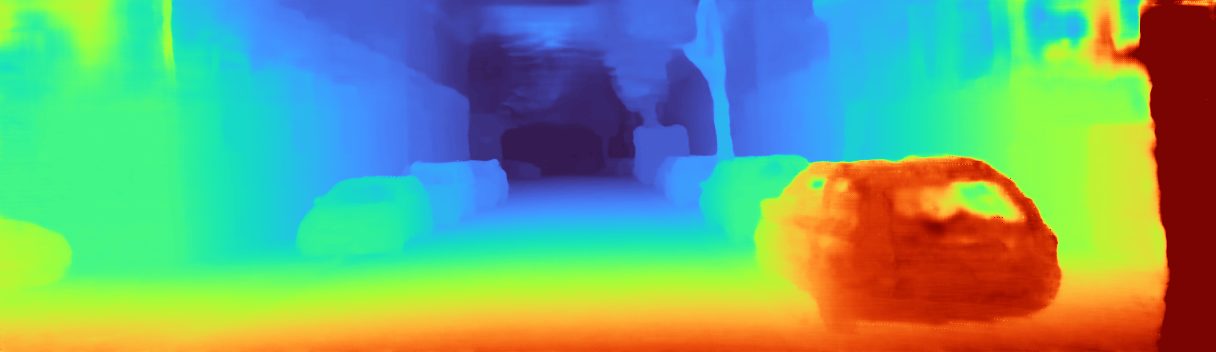}}
     \annotatedFigureBox{0.1,0.454}{0.35,0.974}{B}{0.1,0.454}%bl
     \annotatedFigureBox{0.56,0.67}{0.67,0.98}{C}{0.56,0.7}%bl
     \annotatedLabel{0, 0}{Recurrent (Ours)}
     \end{annotatedFigure}\\
     \begin{annotatedFigure}
     {\includegraphics[trim={8cm 3cm 8cm 0},clip,width=0.4\textwidth]{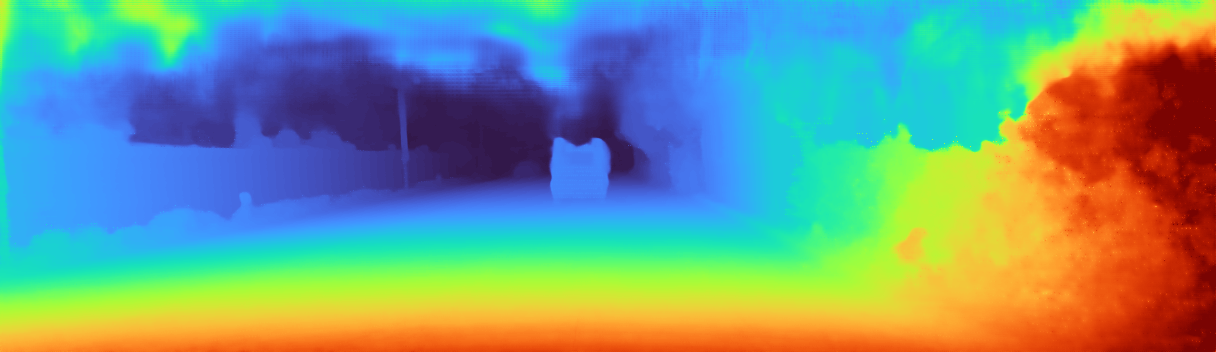}}
     \annotatedFigureBox{0.55,0.5}{0.8,0.98}{A}{0.8,0.5}%bl
     \annotatedLabel{0, 0}{PENet}
     \end{annotatedFigure}%
     \begin{annotatedFigure}
     {\includegraphics[trim={8cm 3cm 8cm 0},clip,width=0.4\textwidth]{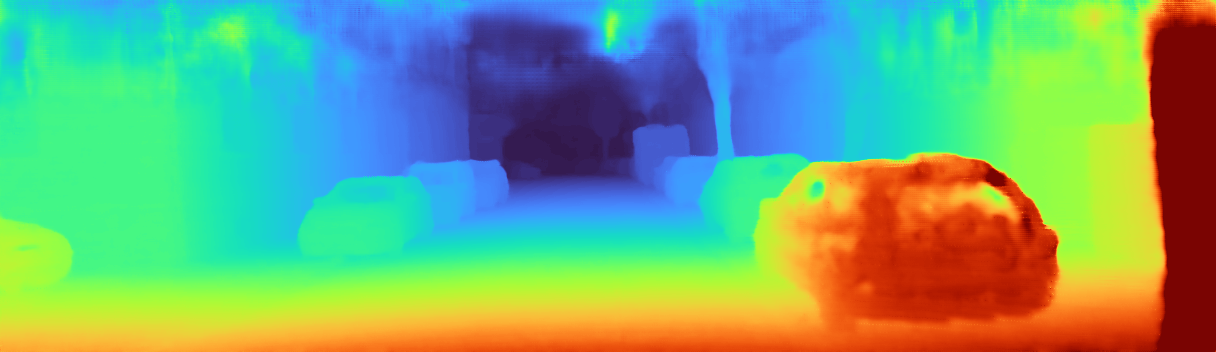}}
     \annotatedFigureBox{0.1,0.454}{0.35,0.974}{B}{0.1,0.454}%bl
     \annotatedFigureBox{0.56,0.67}{0.67,0.98}{C}{0.56,0.7}%bl
     \annotatedLabel{0, 0}{PENet}
     \end{annotatedFigure}\\
     \vspace*{0.3cm}
     \caption{RGB color input and ground truth (top) with recurrent model output (middle) compared to PENet model output (bottom). Notice how the additional temporal information helps especially in regions (A), (B) and (C), where the current timestep sparse lidar input has very few samples.}
     \label{fig:sota_vs_penet1}
\end{figure}

Improvements in such regions are illustrated in \autoref{fig:sota_vs_penet1}, where the increased detail is especially clear on street signs, buildings and trees. These regions contain few sparse depth lidar samples in the current timestep and thus the predicted depth is much more detailed when the model is allowed to warp and accumulate depth samples temporally.

\begin{figure}[!htb]
    \setlength{\lineskip}{0pt}
    \centering
    \begin{annotatedFigure}
    {\includegraphics[height=2cm]{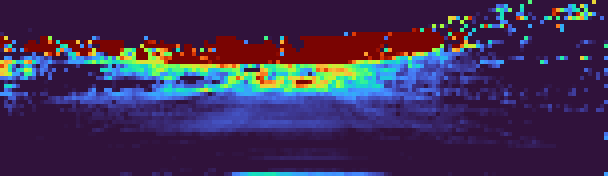}}
    \annotatedLabel{0, 0}{Recurrent (Ours) better}
    \annotatedFigureBox{0.07,0.45}{0.4,0.65}{A}{0.07,0.45}%bl
    \annotatedFigureBox{0.72,0.02}{0.995,0.98}{B}{0.72,0.08}%bl
    \end{annotatedFigure}\\
    \begin{annotatedFigure}
    {\includegraphics[height=2cm]{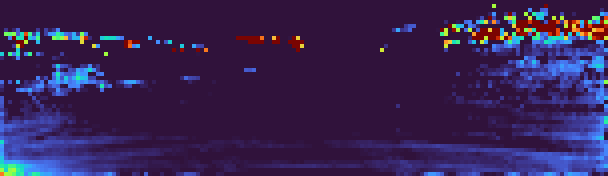}}
    \annotatedLabel{0, 0}{PENet better}
    \annotatedFigureBox{0.07,0.45}{0.4,0.65}{A}{0.07,0.45}%bl
    \annotatedFigureBox{0.72,0.02}{0.995,0.98}{B}{0.72,0.08}%bl
    \end{annotatedFigure}\\
    \vspace*{0.3cm}
     \caption{Difference of average errors for 8 by 8 pixel blocks calculated over the KITTI depth completion validation set. Illustrated using the Turbo colormap \cite{turbo}, where red denotes high values and dark blue low values. Our recurrent model is generally better at mid- to long-range prediction, while matching the accuracy of PENet at short ranges. Regions (A) and (B) highlight the limitations of solutions based on warping, caused by moving objects and small compounding errors.\vspace*{-0.3cm}}
     \label{fig:left_diffs}
\end{figure}

\autoref{fig:left_diffs} illustrates the average difference of errors for 8 by 8 pixel blocks calculated over the KITTI depth completion validation set. The top image illustrates the pixels where our recurrent model has on average lower error than PENet and vice versa for the bottom image. Notice how our recurrent model is more accurate on mid- and long-range depth completion and in regions containing less lidar samples such as the upper parts of the image.

Region (A) of \autoref{fig:left_diffs} contains an interesting phenomenon caused by left-hand traffic, as the vehicles moving to the opposite direction are difficult for both models to estimate. In this scenario, the warped depth is incorrect as warping cannot account for the movement of the other vehicles. We further analyzed this region by calculating the validation metrics only in this region and only for those pixels that are segmented as vehicles by a segmentation network based on \cite{fpn}. While the accuracy improvement on vehicles moving to the opposite direction is not as substantial as the average improvement (approx. 50 RMSE), it is still noteworthy at approximately 25 RMSE.

Notice that the improvement in accuracy of our model near the edges of the \autoref{fig:left_diffs} is below the average improvement. This is especially visible in region (B) where the errors in the intrinsics matrix $K$ and the errors in the measurements between the lidar sensor and right camera are compounded. The warping procedure is especially sensitive to errors in the intrinsics and pose matrices when warping depth samples near the edges, as even small errors can change the location of a warped depth sample by several pixels. \citet{https://doi.org/10.48550/arxiv.2109.03462} have proven that the default calibration parameters of the KITTI dataset are not optimal and that better calibration parameters can significantly increase the accuracy of visual odometry algorithms.

\begin{figure}[!htb]
  \centering
  %\fbox{\rule{0pt}{0.5in} \rule{0.9\linewidth}{0pt}}
  \includegraphics[height=2.5cm]{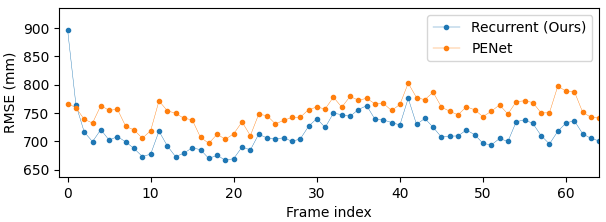}
  \vspace{0.1cm}
   \caption{Average RMSE per frame (from beginning of sequence) on the KITTI depth completion validation set, reported with our recurrent model and the PENet.\vspace*{-0.0cm}}
   \label{fig:timestep}
\end{figure}

The error of our recurrent model decreases quickly during the first frames of
the accumulation. While in the first frame our method is on
average 100 RMSE worse than the PENet baseline, the error
is already 50 RMSE better starting from the third frame.
\autoref{fig:timestep} illustrates the average RMSE per frame compared
to PENet. Note that in a real-world setting, the first frames
can most likely be discarded. For the evaluation, we include
all frames (also the first ones).

\section{Conclusion}

Generating a dense depth map (100\% image coverage) from a sparse depth map with only 6\% coverage seems like a very difficult task. However, by utilizing additional information such as the RGB camera image, the task becomes feasible as previous work has shown.

In this work, we demonstrated a natural extension to this idea: By using a full sequence of consecutive frames with both sparse lidar measurements and camera images as an input to a recurrent neural network, we are able to surpass the current state of the art results on the KITTI validation set.
To this end, we evaluated different options in an ablation study and found that warping the previous predicted depth as an input to the next iteration works best.

In regions without ground truth, we expect the improvement to be even larger. However, this effect is not easily measurable. To verify this assumption, one could manually generate a small test set with dense depth by fitting 3D models into the scene. Another approach would be synthetic data generation.

For the offline use case, we believe accuracy could be further improved by making use of not only past but also future frames. This could be implemented using bidirectional recurrent neural networks \cite{brnn} or via models that explicitly use multiple frames \cite{Kaskela2023}. Another possible future line of work could be to investigate the impact of utilizing more than one former depth frame, as such information would be especially helpful in handling disoccluded regions.

Furthermore, we found our approach to benefit from accurate car ego motion (poses), so combining this approach with leading approaches from the KITTI Visual Odometry challenge \cite{Geiger2012CVPR}, might improve the results further. Integrating a learned approach for pose estimation would also robustify our model against errors originating from the GPS and IMU sensors and allow us to apply our model to indoor depth completion datasets \cite{Silberman:ECCV12,wong2020unsupervised}.

\textbf{Acknowledgments} We thank Jaakko Lehtinen for supervising this project and Pekka Jänis for discussion and reviewing early drafts.

\clearpage
\bibliography{bmvc_final}
\end{document}